\documentclass[11pt]{article}

\usepackage{acl}

\usepackage{times}
\usepackage{latexsym}
\usepackage[T1]{fontenc}
\usepackage[utf8]{inputenc}
\usepackage{microtype}
\usepackage{inconsolata}
\usepackage{graphicx}

\usepackage{booktabs}
\usepackage{multirow}
\usepackage{amsmath}
\usepackage{amssymb}
\usepackage{bm}
\usepackage{enumitem}
\usepackage{xspace}
\usepackage{siunitx}
\usepackage{makecell}
\usepackage{pifont}

\usepackage{tabularx}
\usepackage{array}

\newcolumntype{P}[1]{>{\centering\arraybackslash}m{#1}}

\title{PCoMoE: Shifting MoE Inference from Monolithic Expert Selection to Fine-Grained Path Composition}

\author{
Ziyan Gan$^{1,*}$,
Fangxin Liu$^{1,*,\dagger}$,
Chenyang Guan$^{1}$,
Junjie Wang$^{1}$,
Ning Yang$^{1}$, \\ \bfseries
Haomin Li$^{1}$,
Xiang Li$^{2}$,
Siran Yang$^{2}$,
Jiamang Wang$^{2}$,
Lin Qu$^{2}$, \\ \bfseries
Zongwu Wang$^{1}$,
Li Jiang$^{1,\dagger}$,
Haibing Guan$^{1}$ \\[1mm]
$^{1}$Shanghai Jiao Tong University, China
\qquad
$^{2}$Alibaba Group, China \\[1mm]
 $^{*}$Equal contribution \quad $^{\dagger}$Corresponding authors \\[0.5mm]
 \{Aquarius.o, liufangxin, ljiang\_cs, hbguan\}@sjtu.edu.cn
}

\begin{document}
\maketitle

\begin{abstract}
Mixture-of-Experts (MoE) architectures scale Large Language Model (LLM) capacity efficiently by activating a sparse subset of experts per token. However, modern MoE inference remains heavily constrained by the rigid, whole-expert abstraction. Existing frameworks manage, schedule, or prune experts as atomic execution units, which fixes the optimization boundary too early and leaves fine-grained intra-expert computational redundancy underexplored. In this work, we present \textbf{PCoMoE}, a path-compositional execution framework that shifts MoE inference from coarse-grained expert selection to fine-grained path composition. PCoMoE incorporates a path-level formulation of expert computation, a compatibility-aware layer-wise pruning strategy to suppress low-value path combinations, and a hardware-friendly execution engine to exploit reusable sub-expert structures under strictly bounded overheads. Experimental results demonstrate that PCoMoE achieves up to a 1.31$\times$ end-to-end inference speedup while enhancing model accuracy by 10\%. The code is available at
\url{https://github.com/gzyyy0/PCoMoE}
\end{abstract}

\section{Introduction}

Recent advances in Large Language Models (LLMs) demonstrate that scaling parameter capacity directly drives model capability~\cite{kaplan2020scaling,brown2020language,hoffmann2022training}. Among existing scaling paradigms, the Mixture-of-Experts (MoE) architecture has emerged as a dominant design by decoupling total parameter count from per-token activated computation. By executing only a sparse subset of experts for each input token, MoE models scale model capacity without proportionally increasing training and inference computational costs~\cite{shazeer2017outrageously,lepikhin2020gshard,fedus2022switch,du2022glam}. Consequently, MoE has become a cornerstone for modern high-throughput LLM serving.

\begin{figure}[t]
    \centering
    \includegraphics[width=\columnwidth]{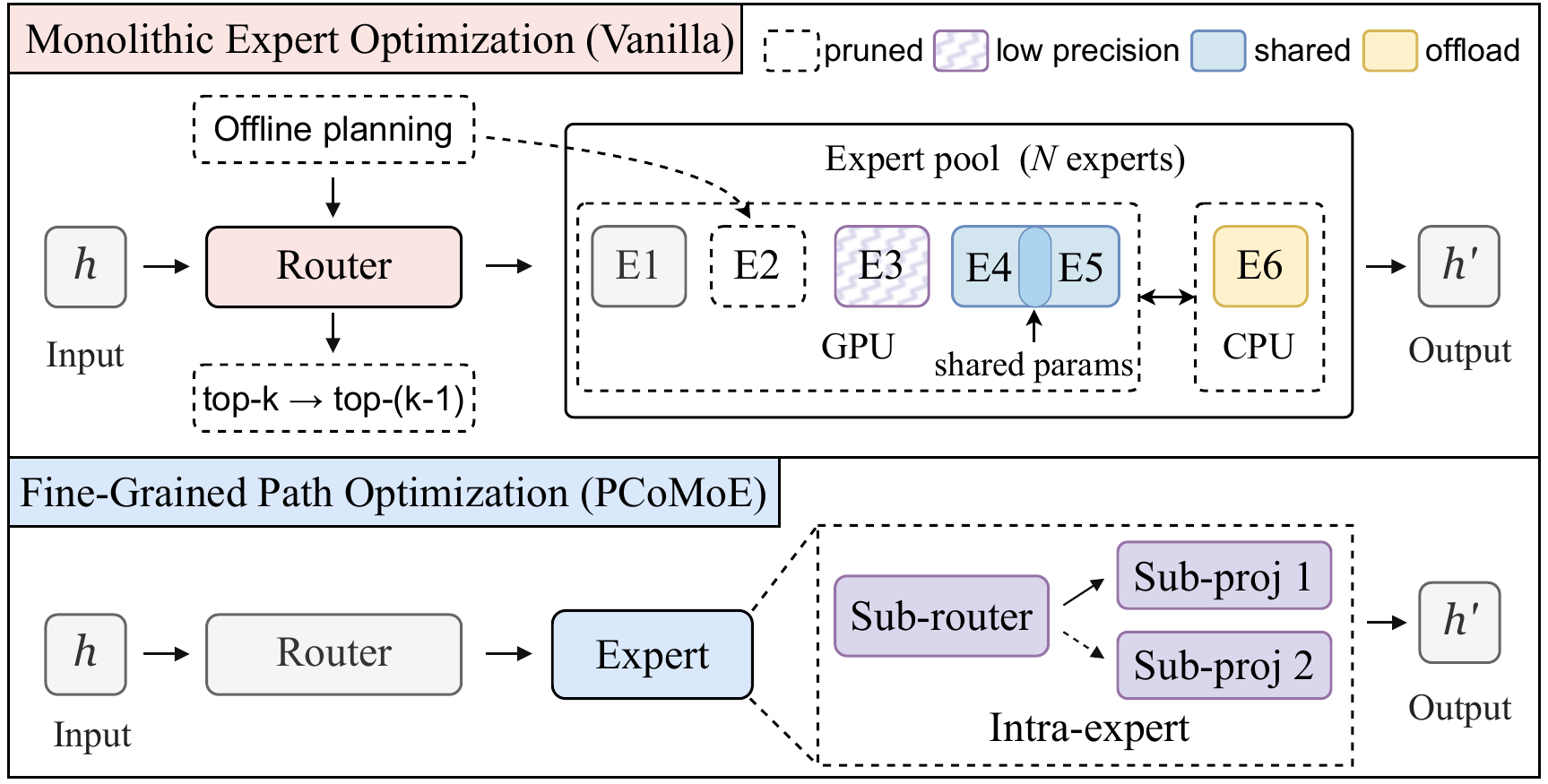}
    \caption{Inter-expert vs. intra-expert optimization in MoE inference. Inter-expert approaches operate on routing decisions and complete experts, whereas intra-expert optimization exposes sub-structures within each expert.}
    \label{fig:intro_expert_centric}
    \vspace{-0.6cm}
\end{figure}

However, theoretical computational sparsity does not automatically translate into end-to-end system efficiency~\cite{liu2021improving}. MoE inference introduces severe dynamic execution overheads, where irregular token-to-expert routing patterns vary drastically across different tokens and layers~\cite{xue2024moeinfinity,tang2024hobbit}. Existing optimization frameworks primarily adopt an \emph{inter-expert} perspective to mitigate these overheads. Complementary model-level acceleration techniques further explore adaptive layer skipping~\cite{liu2025aster} and precision-aware computation~\cite{liu2024spark}. As illustrated in Figure~\ref{fig:intro_expert_centric}, these approaches focus on optimizing scheduling, expert caching, offloading, or expert-level skipping and reuse. Specifically, prior work has explored multiple directions, including simplifying expert selection~\cite{zhong2024adapmoe,chitty2025lexi}, caching and offloading experts~\cite{xue2024moeinfinity,tang2024hobbit}, improving expert placement and prefetching~\cite{fang2025fate,liu2026earth}, and enabling expert-level sharing and reuse~\cite{xshare2026,tan2025rexmoe}. While effective at the macro level, these methods treat each expert as an atomic, indivisible execution unit. This coarse-grained, expert-centric abstraction implicitly assumes that expert computation can only be managed or pruned as a whole, fixing the optimization boundary too early and overlooking the fine-grained computational redundancy buried inside individual experts.

We argue that treating MoE experts as computationally atomic restricts the optimization space and leads to suboptimal efficiency-quality trade-offs. An MoE expert typically consists of structured Feed-Forward Network (FFN) transformations with coupled sub-layers that exhibit distinct semantic roles, computation costs, and quality sensitivities. Restricting optimizations to the inter-expert boundary forces a binary choice between executing or skipping an entire expert, thereby introducing severe representation bottlenecks or unnecessary computation. Breaking this monolithic boundary allows systems to expose finer-grained opportunities for intra-expert reuse and reorganization, which is essential to jointly eliminate computational redundancy and maintain model accuracy.

To exploit these fine-grained opportunities, we present \textbf{PCoMoE}, a path-compositional execution framework that optimizes MoE inference via intra-expert sub-transformation composition. PCoMoE decomposes monolithic experts into composable internal sub-structures, transforming standard expert-level selection into path-level composition over an expanded path space. Instead of relying on rigid expert boundaries, PCoMoE adaptively reorganizes internal expert computations to maximize inference efficiency. To regulate the expanded path space without introducing execution chaos, PCoMoE incorporates compatibility-aware path routing and layer-wise pruning to suppress low-value combinations based on layer-specific sensitivity. Furthermore, we co-design a hardware-friendly path execution engine that groups dispatches by source and shares expansion-side operations, converting path-level flexibility into real end-to-end speedups under strictly bounded system overheads.
Our main contributions are threefold. \textbf{(1) Path-Level Composition:} We reformulate MoE inference from vanilla monolithic selection to fine-grained sub-transformation composition, unlocking latent intra-expert execution trajectories while preserving baseline vanilla pathways. \textbf{(2) Compatibility-Aware Gating:} We introduce structured compatibility gating with iterative structural pruning to suppress low-value off-diagonal routes, minimizing combinatorial scheduling overheads while guaranteeing representation fidelity. \textbf{(3) Hardware-Efficient Runtime:} We develop an execution pipeline mapping active compositional paths directly to source-grouped compute clusters, translating algorithmic flexibility into deterministic acceleration. Comprehensive evaluations show that PCoMoE achieves up to a 1.31$\times$ serving speedup while consistently outperforming vanilla MoE baselines across downstream tasks by up to 10\% in accuracy.

\section{Background}
\begin{figure}[t]
    \centering
    \includegraphics[width=\columnwidth]{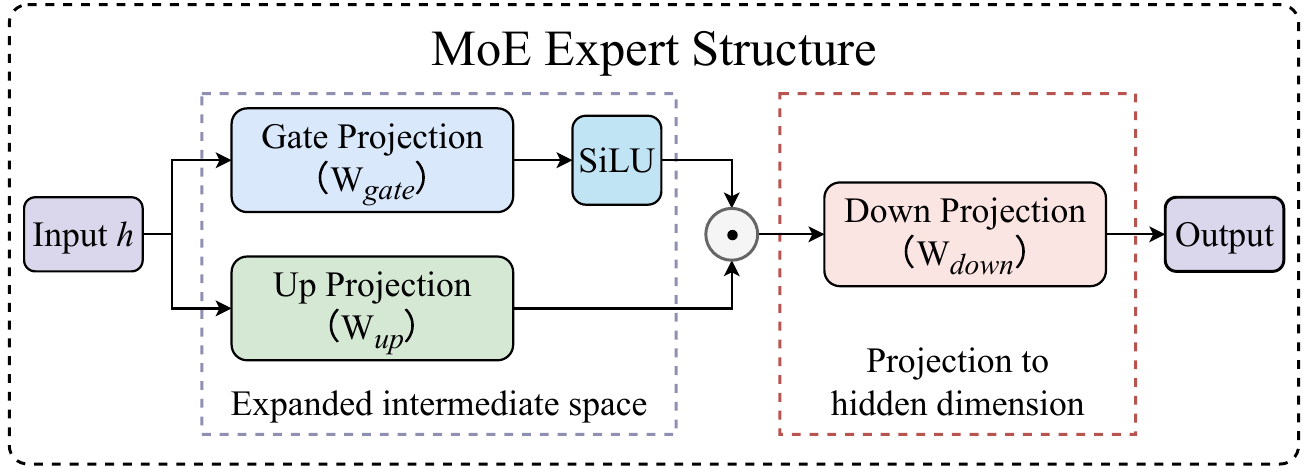}
    \caption{Representative internal structure of an MoE expert, illustrated with a SwiGLU-style gated FFN.}
    \label{fig:swiglu_expert}
    \vspace{-0.5cm}
\end{figure}

A vanilla MoE layer consists of a router and a set of sparse feed-forward experts~\cite{shazeer2017outrageously,lepikhin2020gshard,fedus2022switch}. Given the hidden representation $h$ of an input token, the router computes routing scores to select the top-$k$ experts for execution. Denoting the selected expert set by $\mathcal{E}(h)$, the output of the MoE layer is formulated as
\[
y=\sum_{e\in\mathcal{E}(h)} \alpha_e\,F_e(h),
\]
where $\alpha_e$ is the routing weight assigned to expert $e$, and $F_e(\cdot)$ denotes the non-linear transformation implemented by expert $e$. At runtime, tokens are dispatched according to routing decisions, grouped by expert for batched execution, and scattered back before weighted aggregation. This mechanism dictates that a vanilla MoE layer operates on a rigid dispatch-execute-combine pipeline~\cite{lepikhin2020gshard,xue2024moeinfinity}. Consequently, MoE inference introduces highly input-dependent execution footprints, where activated expert profiles vary dynamically across tokens and layers.

Although vanilla MoE is traditionally managed at the expert selection level, each individual expert is a structured feed-forward module rather than an atomic computational primitive. Modern MoE models predominantly employ the SwiGLU-style gated FFN architecture~\cite{shazeer2020glu} as their core expert instantiation, a design validated in representative open-source models including Mixtral and Qwen-MoE~\cite{jiang2024mixtral,qwen15moe2024}. For an input hidden state $h$, the computation of expert $e$ is defined as
\[
F_e(h)=W^{(e)}_{\mathrm{down}}
\Bigl(
\mathrm{SiLU}\!\left(W^{(e)}_{\mathrm{gate}}h\right)
\odot
\left(W^{(e)}_{\mathrm{up}}h\right)
\Bigr),
\]
where $W^{(e)}_{\mathrm{gate}}$ and $W^{(e)}_{\mathrm{up}}$ project the input tensor into an expanded intermediate space, $\odot$ denotes element-wise multiplication, and $W^{(e)}_{\mathrm{down}}$ maps the fused activations back to the hidden dimension.

Figure~\ref{fig:swiglu_expert} illustrates this intra-expert computational layout. The gate and up branches execute parallel matrix multiplications on the expansion side, followed by element-wise activation fusion and a final down projection. Mechanistically, these parallel branches serve distinct algorithmic roles: the gate branch modulates activation thresholds, the up branch extracts expanded feature representations, and the down branch aggregates these features back into the model dimension. This inner layout implies that the computational workload within an MoE expert can be reorganized at a finer granularity than the atomic whole-expert interface enforced by vanilla routing.

In actual hardware deployments, end-to-end MoE execution latency is governed by system-level orchestrations beyond raw arithmetic FLOPs. It encompasses routing overhead, token dispatch, expert-wise packing, batched kernel launches, result scattering, and weighted tensor aggregation~\cite{xue2024moeinfinity,tang2024hobbit}. Existing hardware-aware optimizations generally target two distinct granularities: minimizing the expert computation exposed by the router through pruning or skipping, or maximizing the execution efficiency of selected experts via batched routing, sharing, caching, or prefetching. Despite mitigating different system bottlenecks, these frameworks consistently preserve the whole-expert abstraction as their basic execution unit. This rigid boundary artificially restricts the optimization space, concealing critical opportunities for structural reorganization and cross-expert computation reuse within individual sub-expert layers.

\section{Motivation}
\begin{figure}[t]
    \centering
    \includegraphics[width=\columnwidth]{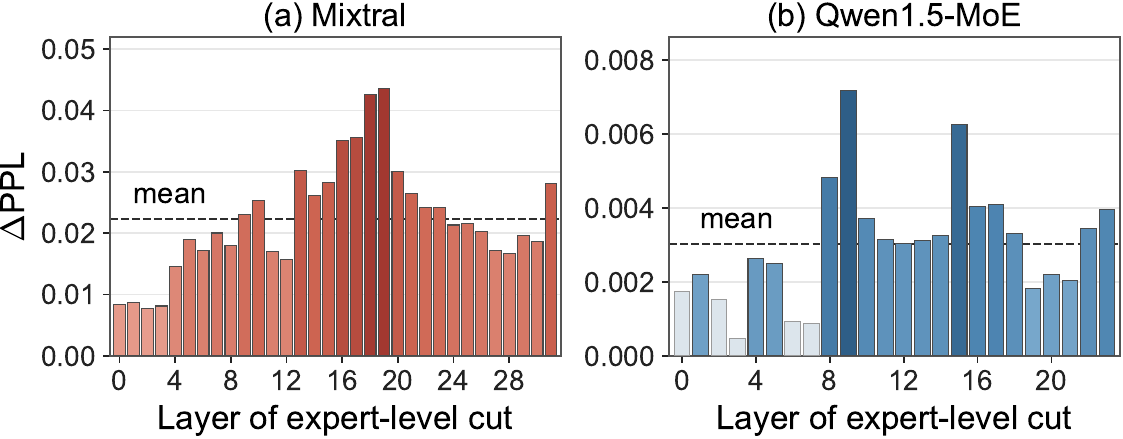}
    \caption{Quality impact of single-layer whole-expert reduction. Each bar reports the $\Delta$PPL after changing top-$k$ to top-$(k-1)$ routing at one MoE layer; darker bars indicate larger values.}
    \label{fig:motivation_single_cut}
    \vspace{-0.5cm}
\end{figure}

Vanilla MoE inference restricts routing to monolithic expert boundaries. While computationally convenient, this coarse abstraction overlooks the structural misalignment between sparsity-induced redundancy and indivisible expert modules.

\subsection{Inefficiency of Coarse-Grained Routing}
\label{subsec:coarse}

Traditional MoE acceleration relies on eliminating entire expert executions. Since the vanilla MoE interface exposes computation via top-$k$ routing, a baseline replaces top-$k$ routing with top-$(k-1)$ routing to decrease the number of activated experts per token. We sweep individual layers to profile the perplexity impact of this single-layer pruning. 

As shown in Figure~\ref{fig:motivation_single_cut}, imposing the same expert-level pruning across different layers yields highly divergent quality degradation. Certain layers exhibit minimal perplexity increases, whereas others incur substantial degradation that exceeds the average trend, proving that whole-expert reduction is neither uniformly safe nor uniformly detrimental.
This variation demonstrates that inter-expert redundancy is highly non-homogeneous and layer-dependent. While specific layers tolerate complete expert deletion, this local redundancy cannot be generalized into a fixed global pruning rule. Multi-layer reduction compounds these errors.

\begin{figure}[t]
    \centering
    \includegraphics[width=\columnwidth]{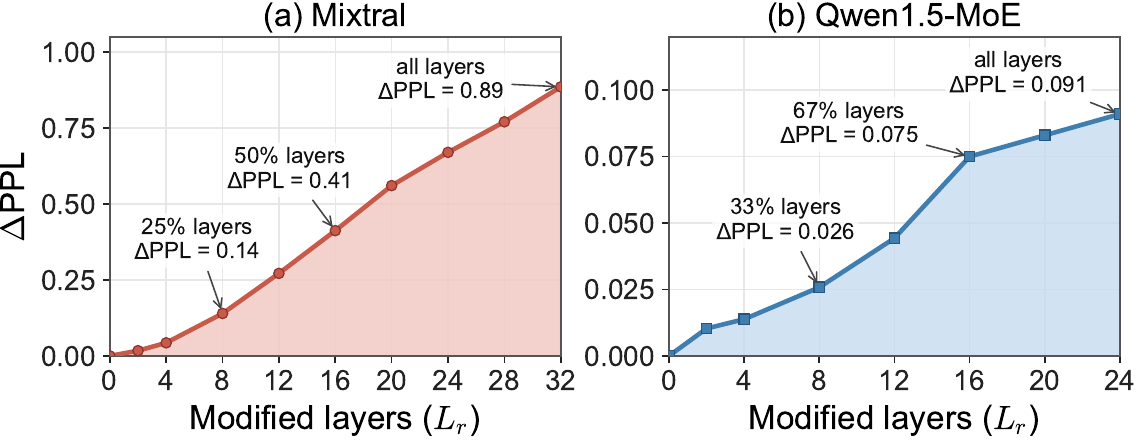}
    \caption{Cumulative quality impact of reducing one activated expert across multiple MoE layers.}
    \label{fig:motivation_accumulated_loss}
    \vspace{-0.5cm}
\end{figure}

As illustrated in Figure~\ref{fig:motivation_accumulated_loss}, local errors compound catastrophically across deep networks when expert reduction is extended globally. Although the absolute degradation scales differently across models, the underlying trend remains consistent: locally tolerable approximations aggregate into severe end-to-end quality drops. Consequently, repeated whole-expert reduction rapidly depletes safe redundancy, proving that monolithic expert boundaries are too coarse for optimal efficiency-quality trade-offs.

The central challenge is therefore not merely adjusting the number of reduced experts, but breaking the whole-expert abstraction to exploit finer computational granularities.

\begin{figure}[t]
    \centering
    \includegraphics[width=\columnwidth]{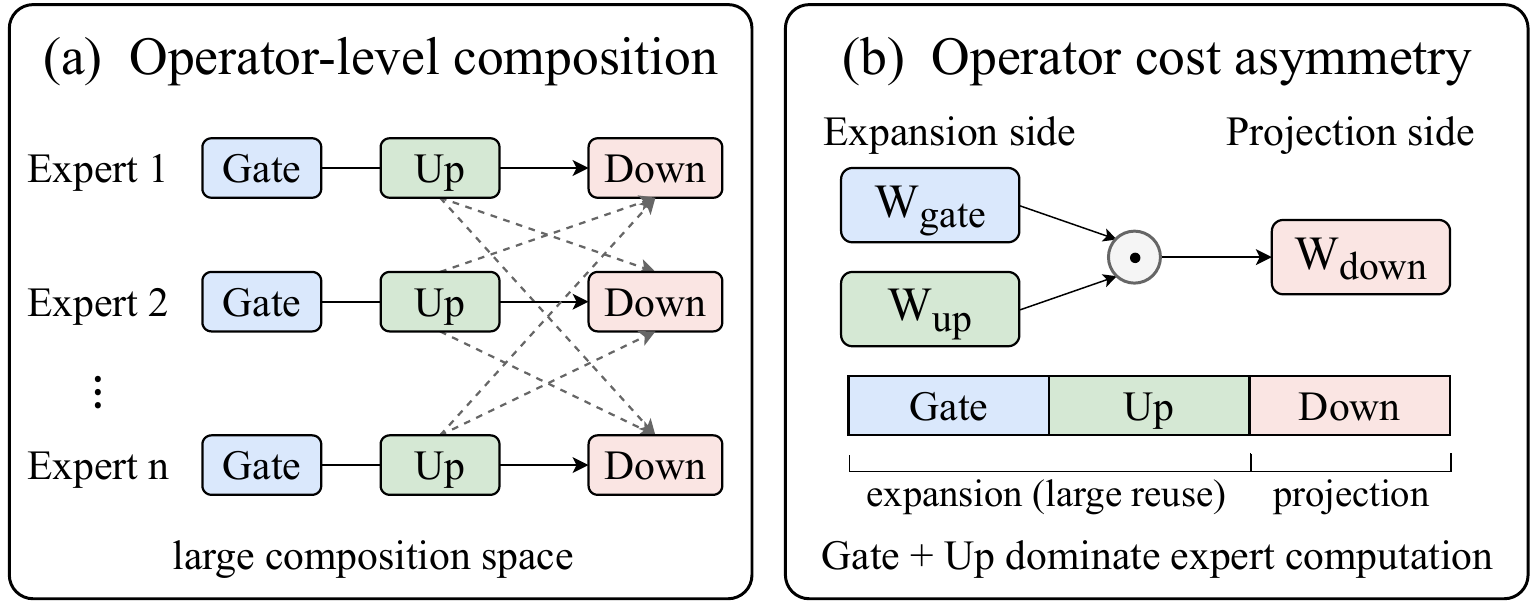}
    \caption{Intra-expert composition and cost asymmetry. (a) Separately composing Gate, Up, and Down expands the routing decisions. (b) Grouping Gate and Up as the expansion side, and Down as the projection side, exposes a structured reuse boundary with asymmetric computational costs.}
    \label{fig:motivation_operator_asymmetry}
    \vspace{-0.5cm}
\end{figure}

\begin{figure*}[t]
    \vspace{-0.5cm}
    \centering
    \includegraphics[width=0.9\textwidth]{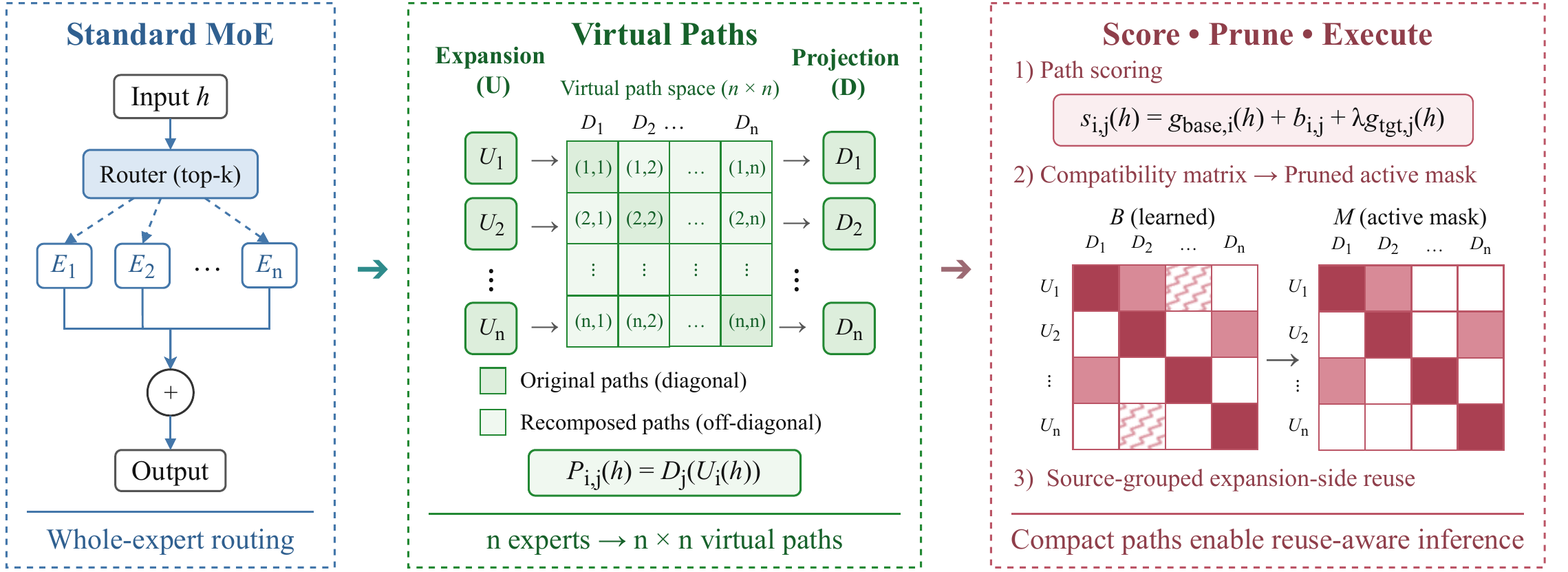}
    \caption{Overview of PCoMoE. 
    PCoMoE decomposes each expert into expansion-side and projection-side operators, forming an expanded path space over their cross-compositions. To make this space practical, PCoMoE scores compositional paths via learned compatibility, prunes low-value paths into a compact active set, and executes the remaining paths through source-grouped expansion-side reuse.}
    \label{fig:vemoe_overview}
    \vspace{-0.5cm}
\end{figure*}

\subsection{Intra-Expert Operator Asymmetry}

To bypass these bottlenecks, we analyze the internal computational structures of individual experts.  A vanilla SwiGLU expert maps inputs through parallel gate and up projections on the expansion side (comprising two-thirds of total expert FLOPs) before a final down projection back to the hidden dimension. Directly treating these operators as independent, unconstrained composition units (Figure~\ref{fig:motivation_operator_asymmetry}(a)) introduces prohibitive control-flow overheads. Instead, the intrinsic expansion-projection boundary (Figure~\ref{fig:motivation_operator_asymmetry}(b)) defines a structured middle ground. In a top-2 Mixtral architecture, reusing the expansion side eliminates approximately one-third of total expert computation, whereas reusing the projection side saves only one-sixth, concentrating potential efficiency gains on the expansion side.

Profiling operator substitution on Mixtral confirms that this compute-dense expansion side also exhibits superior structural tolerance for representation sharing. Substituting the gate and up projections with an alternative source expert yields lower representation error in 23 out of 32 layers under optimal independent pairing. Under a stricter constraint where all experts within a layer share a single replacement source, 25 out of 32 layers still favor expansion-side substitution. This dual computational and structural asymmetry demonstrates that the expansion-projection boundary serves as an optimal abstraction, avoiding the accuracy drops of whole-expert dropping while circumscribing the irregular execution overheads of arbitrary operator splitting.

However, translating this theoretical boundary into serving gains introduces three critical co-design challenges: (1) the path search space scales combinatorially compared to vanilla expert selection, (2) additional control logic or dynamic indexing can easily offset computational savings under strict latency constraints, and (3) compositional stability varies heavily across layers, threatening feature fidelity. Addressing these hardware-software trade-offs requires an execution framework that simultaneously regulates the path composition space, minimizes runtime dispatch overheads, and guarantees model quality.

\section{Design}
\label{sec:design}

Figure~\ref{fig:vemoe_overview} illustrates the PCoMoE overview, which bypasses the monolithic constraints of vanilla MoE inference via fine-grained sub-transformation composition. While preserving the vanilla routing-compatible interface at the layer boundary, PCoMoE decouples each internal expert into standalone expansion-side and projection-side operators. For an MoE layer with $n$ experts, cross-composing these operators expands the optimization landscape into an $n \times n$ compositional path space. Within this matrix, diagonal trajectories maintain the original expert computations to anchor baseline stability, whereas off-diagonal paths represent newly synthesized execution routes that introduce zero parameter or storage overheads.

To prevent this expanded space from inducing combinatorial scheduling overheads, PCoMoE actively filters candidate trajectories via a compatibility-guided pruning mechanism. The framework assigns each path a composite score that balances vanilla routing priors with learned source-to-target compatibility. This scoring function actively suppresses low-value off-diagonal trajectories, confining the search space to a compact, high-fidelity set of active execution paths.

The remaining active paths map directly onto a hardware-efficient pipeline that eliminates redundant kernel launches through source-grouped computation reuse. Instead of executing each path in isolation, PCoMoE aggregates tokens sharing identical expansion-side operators, allowing them to reuse a single compute footprint before dynamic dispatching to their respective projection-side destinations. This hardware-software co-design seamlessly translates algorithmic path-level flexibility into end-to-end serving acceleration under strictly bounded runtime overheads.

\subsection{Compositional Path Formulation}
\label{sec:virtual_path_formulation}
PCoMoE redefines the computational granularity of the MoE layer by decoupling SwiGLU experts along their structural asymmetry boundaries. The expansion-side operator aggregates the compute-dominant gate and up projections, while the projection-side operator maps intermediate activations back to the hidden dimension. Formulating their ordered combination as a compositional path concentrates optimization capabilities precisely where computational reuse potential is maximized, shifting the execution paradigm from monolithic expert selection to flexible path composition.

For an MoE layer containing $n$ physical experts, the computation of an individual expert $e$ is decoupled into an expansion-side function $U_e(h)$ and a projection-side function $D_e(z)$. We express the expansion side as
\[
U_e(h)=\mathrm{SiLU}\!\left(W^{(e)}_{\mathrm{gate}}h\right) \odot \left(W^{(e)}_{\mathrm{up}}h\right),
\]
and the projection side as
\[
D_e(z)=W^{(e)}_{\mathrm{down}}z,
\]
which reconstructs the original vanilla expert transformation via
\[
F_e(h)=D_e(U_e(h)).
\]

We define the pools of reusable expansion and projection operators across the layer as $\mathcal{U}=\{U_1,\ldots,U_n\}$ and $\mathcal{D}=\{D_1,\ldots,D_n\}$, respectively. A compositional path $P_{i,j}(h)$ represents the ordered composition of an expansion source $i$ and a projection target $j$:
\[
P_{i,j}(h)=D_j(U_i(h)), \qquad i,j\in\{1,\ldots,n\}.
\]
This formulation expands the available computational trajectories from $n$ physical modules to an $n^2$ compositional space. The diagonal path $P_{i,i}$ represents the original vanilla expert $F_i$, while an off-diagonal path $P_{i,j}$ ($i \neq j$) synthesizes a new computational trajectory combining the expansion side of expert $i$ with the projection side of expert $j$.

This optimization landscape is purely architectural and introduces zero parameter or storage overheads. All routes are dynamically synthesized by recombining existing physical operators without creating new weights. Restricting execution to diagonal paths naturally reduces PCoMoE back to the vanilla MoE layer, whereas off-diagonal paths serve as selectively activated candidates to maximize efficiency-quality trade-offs.

Consequently, the MoE layer output generalizes from expert-level routing to path-level aggregation. Let $\Pi(h) \subseteq \{1,\ldots,n\} \times \{1,\ldots,n\}$ denote the active compositional paths selected for a token hidden state $h$, and let $\beta_{i,j}(h)$ represent the corresponding path weight. The PCoMoE layer output is formulated as
\[
y=\sum_{(i,j)\in\Pi(h)} \beta_{i,j}(h) P_{i,j}(h).
\]
The vanilla MoE layer remains a strict subset of this generalized formulation, where setting $\Pi(h)=\{(e,e) \mid e\in\mathcal{E}(h)\}$ and $\beta_{e,e}(h)=\alpha_e(h)$ exactly recovers the original weighted expert summation. Although this formulation unlocks a fine-grained optimization landscape, evaluating the entire $n^2$ space introduces prohibitive computational overheads. PCoMoE treats this matrix strictly as a candidate space, using a compatibility-driven pruning mechanism to isolate a compact, high-value active path set for execution.

\subsection{Compatibility-Aware Path Routing}
\label{sec:path_routing_pruning}
\begin{figure}[t]
    \centering
    \includegraphics[width=\columnwidth]{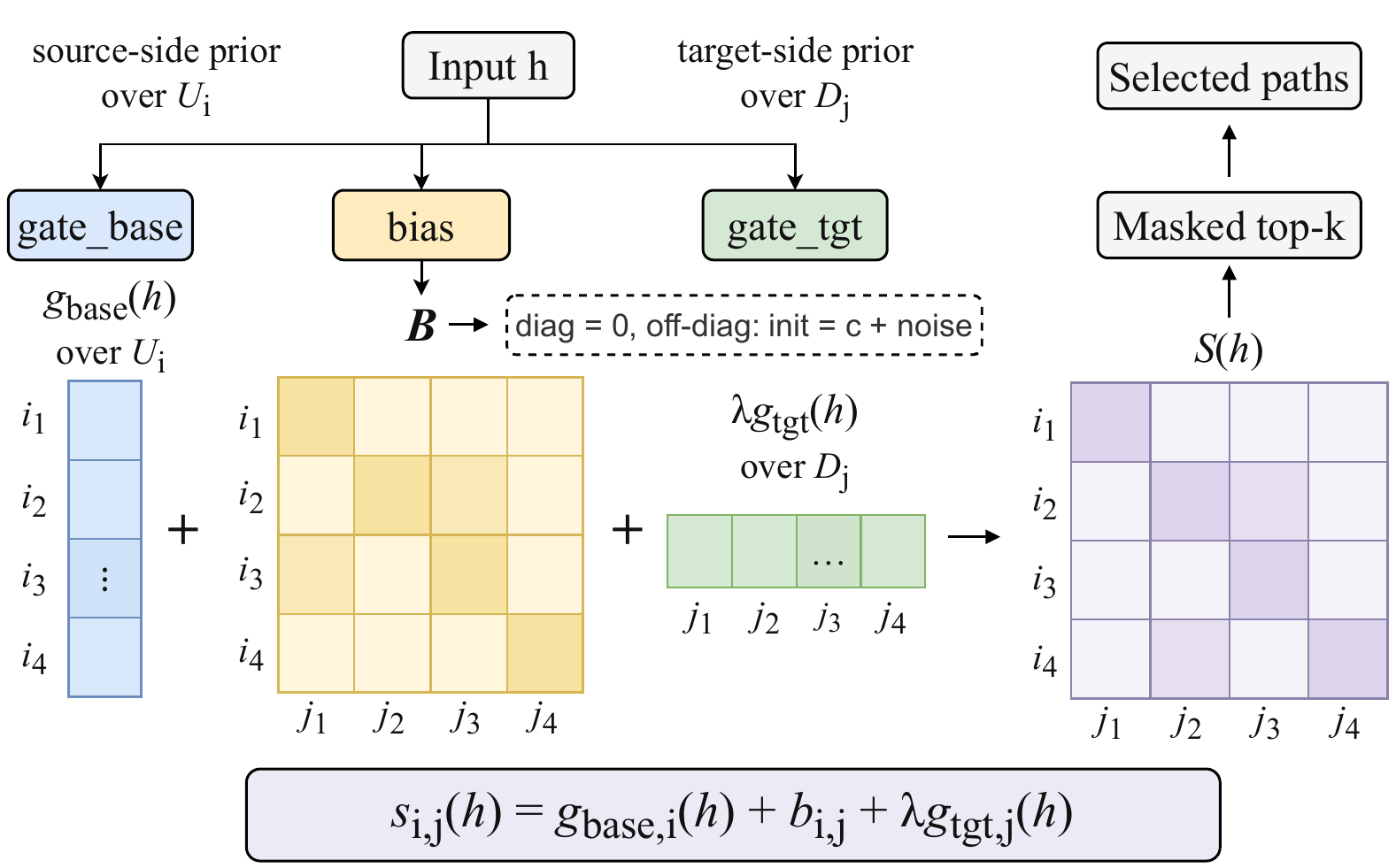}
    \caption{Compatibility-aware path gate for scoring source-to-target compositional paths.}
    \label{fig:path_gate_pruning}
    \vspace{-0.5cm}
\end{figure}

The compositional path formulation expands the architectural search landscape into an $n \times n$ routing matrix. To suppress noisy off-diagonal recompositions and bound runtime dispatch complexity, PCoMoE introduces a compatibility-aware path gating mechanism (Figure~\ref{fig:path_gate_pruning}) that factors each trajectory score into a source-side routing prior, a target-side routing prior, and a learned pairwise compatibility bias. For a compositional path $P_{i,j}^{(\ell)}$ in layer $\ell$, the gating function computes the composite routing score as
\[
s_{i,j}^{(\ell)}(h) = g_{\mathrm{base},i}^{(\ell)}(h) + b_{i,j}^{(\ell)} + \lambda g_{\mathrm{tgt},j}^{(\ell)}(h),
\]
where $g_{\mathrm{base}}^{(\ell)}(h)\in\mathbb{R}^{n}$ denotes the base expert-routing signal, $B^{(\ell)}=[b_{i,j}^{(\ell)}]\in\mathbb{R}^{n\times n}$ represents the learned compatibility bias, $g_{\mathrm{tgt}}^{(\ell)}(h)\in\mathbb{R}^{n}$ provides a projection-side prior, and $\lambda$ scales this target-side preference. The base routing signal anchors the optimization to the pretrained vanilla checkpoints by broadcasting across source indices, while the target prior regularizes projection choices. Crucially, the compatibility matrix $B^{(\ell)}$ parameterizes the structural alignment between expansion source $U_i$ and projection target $D_j$, transforming the raw Cartesian product into a filtered execution topology.

PCoMoE initializes this compatibility matrix to strictly prioritize vanilla computational trajectories. Diagonal entries are zero-initialized via $b_{i,i}^{(\ell)}=0$ to safeguard baseline expert behavior, whereas off-diagonal elements receive a negative offset with bounded random noise:
\[
b_{i,j}^{(\ell)}\sim c+\epsilon,\qquad i\neq j,\; c<0.
\]
This negative prior prevents unoptimized off-diagonal trajectories from disrupting early training phases while breaking structural symmetry. During fine-tuning, this bias is dynamically converted into a layer-specific active mask $M^{(\ell)}\in\{0,1\}^{n\times n}$. Rather than applying a post-hoc, one-shot pruning step, PCoMoE iteratively tightens a structural sparsity criterion at progressive training intervals to filter low-value off-diagonal routes while hard-coding $m_{i,i}^{(\ell)}=1$. This continuous regularization allows the remaining routing weights to co-adapt with the contracting active path set, preserving model convergence stability.

The active mask operates independently across layers on routing eligibility rather than on physical parameters, enabling each layer to autonomously converge to its optimal path density based on local feature sensitivity. Because no physical weights are deleted, suppressed paths are merely excluded from the active runtime dispatch graph. During inference, inactive paths are masked out prior to gating, ensuring that the online router evaluates only the optimized, compact set of active trajectories induced by $M^{(\ell)}$, thereby successfully minimizing control-flow overheads.

\begin{table*}[t]
    \vspace{-0.7cm}
    \centering
    \footnotesize
    \caption{Zero-shot accuracy across MoE backbones (higher is better).}
    \label{tab:quality_main}
    \vspace{-0.25cm} 
    \setlength{\tabcolsep}{4.0pt} 
    \renewcommand{\arraystretch}{0.95} 
    \begin{tabular}{@{}cccccccc@{}}
        \toprule
        \textbf{Model} & \textbf{Method} & \textbf{BoolQ} & \textbf{ARC-E} & \textbf{ARC-C} & \textbf{HellaSwag} & \textbf{WinoGrande} & \textbf{Avg.} \\
        \midrule

        \multirow{7}{*}{Mixtral-8x7B}
        & Vanilla & 85.93 & 82.62 & 59.56 & 84.10 & 77.11 & 77.86 \\
        & Vanilla-FT & 86.45 & 83.78 & 61.68 & 83.34 & 76.56 & 78.36 \\
        & MoE-I$^2$ & 82.60 & 78.20 & 52.20 & 61.10 & 71.50 & 69.10 \\
        & MoE-Pruner & 86.00 & 81.90 & 53.30 & 62.30 & 75.50 & 71.80 \\
        & MoEITS & 87.33 & 83.08 & 55.89 & 82.83 & \textbf{80.98} & 78.02 \\
        \cmidrule(lr){2-8}
        & \textbf{PCoMoE} & \textbf{88.38} & \textbf{85.06} & \textbf{62.71} & \textbf{85.46} & 78.22 & \textbf{79.97} \\
        & \emph{$\Delta$ vs Vanilla-FT} & +1.93 & +1.28 & +1.03 & +2.12 & +1.66 & +1.61 \\

        \midrule

        \multirow{7}{*}{Qwen1.5-MoE}
        & Vanilla & 79.72 & 69.28 & 44.20 & 77.25 & 69.06 & 67.90 \\
        & Vanilla-FT & 82.02 & 77.01 & 51.10 & 74.96 & 69.06 & 70.83 \\
        & MoE-I$^2$ & 75.08 & 71.68 & 41.13 & 53.08 & 66.54 & 61.50 \\
        & MoE-Pruner & 69.14 & 52.02 & 29.10 & 42.99 & 59.12 & 50.47 \\
        & MoEITS & 75.20 & 72.30 & 36.01 & 61.27 & 67.46 & 62.45 \\
        \cmidrule(lr){2-8}
        & \textbf{PCoMoE} & \textbf{86.18} & \textbf{80.43} & \textbf{52.30} & \textbf{77.67} & \textbf{71.90} & \textbf{73.70} \\
        & \emph{$\Delta$ vs Vanilla-FT} & +4.16 & +3.42 & +1.20 & +2.71 & +2.84 & +2.87 \\

        \midrule

        \multirow{7}{*}{DeepSeek-V2-Lite}
        & Vanilla & 80.61 & 74.16 & 46.33 & \textbf{77.74} & \textbf{71.27} & 70.02 \\
        & Vanilla-FT & 81.83 & 75.32 & 47.94 & 76.23 & 70.40 & 70.34 \\
        & MoE-I$^2$ & 76.79 & 71.80 & 42.58 & 55.16 & 67.64 & 62.79 \\
        & MoE-Pruner & 76.61 & 71.89 & 40.02 & 50.94 & 67.64 & 61.42 \\
        & MoEITS & 80.03 & 77.61 & 43.77 & 70.80 & 67.72 & 67.99 \\
        \cmidrule(lr){2-8}
        & \textbf{PCoMoE} & \textbf{83.94} & \textbf{79.59} & \textbf{51.02} & 76.66 & 71.19 & \textbf{72.48} \\
        & \emph{$\Delta$ vs Vanilla-FT} & +2.11 & +4.27 & +3.08 & +0.43 & +0.79 & +2.14 \\

        \bottomrule
    \end{tabular}
    \vspace{-0.35cm} 
\end{table*}

\subsection{Hardware-Efficient Path Execution}
\label{sec:system_friendly_execution}

\begin{figure}[t]
    \centering
    \includegraphics[width=\columnwidth]{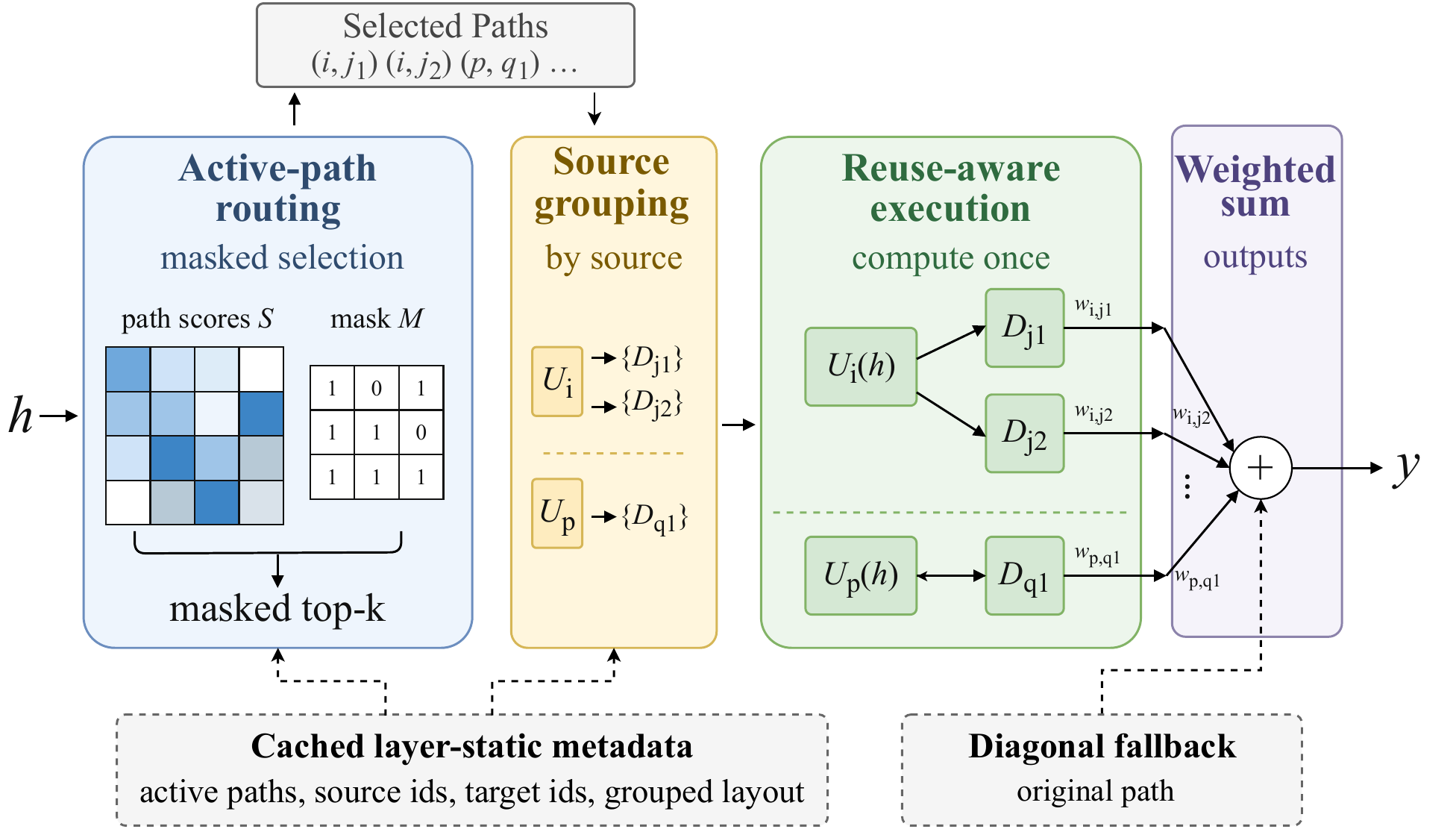}
    \caption{Hardware-efficient inference execution of active compositional paths in PCoMoE.}
    \label{fig:system_execution}
    \vspace{-0.5cm}
\end{figure}

To translate the pruned path space into physical serving gains, PCoMoE decouples its runtime execution into a routing phase and a source-grouped reuse phase (Figure~\ref{fig:system_execution}). 

\textbf{Compositional Path Routing.} 
During inference, the framework isolates active trajectories by masking the $n \times n$ routing matrix with the pruned path set $\mathcal{A}^{(\ell)} = \{(i,j) \mid m_{i,j}^{(\ell)}=1\}$. The router selects the active path subset via
\[
\Pi^{(\ell)}(h) = \mathrm{TopK}_{(i,j)\in \mathcal{A}^{(\ell)}} s_{i,j}^{(\ell)}(h),
\]
collapsing dense evaluation into a sparse search. PCoMoE materializes path indices, source-to-target layouts, and grouped dispatch metadata offline. Based on this cached layout, the fusion optimization combines online filtering, top-$k$ selection, and dispatch preparation into a single lightweight indexing operation, avoiding reconstruction of the full $n \times n$ path space. Because all diagonal paths remain active, the routing graph automatically falls back to vanilla expert execution under low-confidence scenarios, preserving the native MoE interface.

\textbf{Source-Grouped Compute Reuse.} 
Following routing, PCoMoE aggregates active trajectories by their expansion-side sources to eliminate redundant tensor evaluations. This source-grouped dispatch changes the dispatch unit from individual paths $(i,j)$ to source groups, allowing paths that share the same expansion-side operator to be scheduled together. Denoting the unique selected sources as $\mathcal{S}^{(\ell)}(h)=\{i \mid \exists j, (i,j)\in \Pi^{(\ell)}(h)\}$ and their associated projection targets as $\mathcal{T}_{i}^{(\ell)}(h) = \{j \mid (i,j)\in \Pi^{(\ell)}(h)\}$, the final layer output is formulated as:
\[
y = \sum_{i\in\mathcal{S}^{(\ell)}(h)} \sum_{j\in\mathcal{T}_{i}^{(\ell)}(h)} \beta_{i,j}^{(\ell)}(h)\,D_j\left(U_i(h)\right).
\]
By executing each compute-heavy expansion operator $U_i(h)$ exactly once per unique source, expansion evaluations scale with the active source count $|\mathcal{S}^{(\ell)}(h)|$ instead of the total path budget $|\Pi^{(\ell)}(h)|$. This hardware-software co-design converts structural sharing into deterministic latency reductions.

\begin{figure*}[t]
    \vspace{-0.7cm}
    \setlength{\abovecaptionskip}{3pt}
    \setlength{\belowcaptionskip}{3pt}
    \centering
    \includegraphics[width=\textwidth]{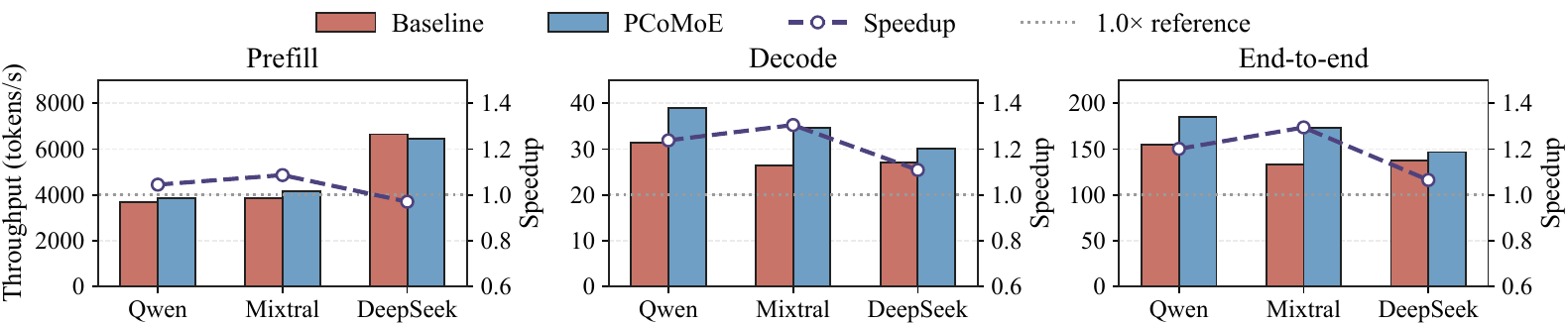}
    \caption{Token throughput and speedup of PCoMoE against vanilla MoE baselines.}
    \label{fig:main_speed}
    \vspace{-0.5cm}
\end{figure*}

\section{Evaluation}
\label{sec:Evaluation}

\subsection{Experimental Setup}
\label{sec:eval_setup}

\paragraph{Implementation.} 
Performance profiling is conducted on a server equipped with NVIDIA H20 GPUs and Intel Xeon 6759P-C CPUs. All evaluations are performed on a single GPU under a unified precision and software environment.

\paragraph{Models.} 
We validate PCoMoE across three MoE models spanning diverse routing granularities: Qwen1.5-MoE-A2.7B ($60$ routed experts, top-$4$ activation)~\cite{qwen15moe2024}, Mixtral-8x7B-v0.1 ($8$ routed experts, top-$2$ activation)~\cite{jiang2024mixtral}, and DeepSeek-V2-Lite ($64$ routed experts, top-$6$ activation)~\cite{deepseek2024v2}.


\paragraph{Fine-tuning.}
We use a 25K-example mixture of Alpaca and SQuAD with AdamW. Vanilla-FT and PCoMoE use the same LoRA configuration for gating adaptation, with rank 16 and a batch size of 32, while all expert weights remain frozen. PCoMoE additionally learns path compatibility biases for compositional routing and progressive pruning.

\paragraph{Datasets.} 
Downstream model quality is verified on five benchmarks: BoolQ~\cite{clark2019boolq}, ARC-Easy, ARC-Challenge~\cite{clark2018arc}, HellaSwag~\cite{zellers2019hellaswag}, and WinoGrande~\cite{sakaguchi2019winogrande}. We report standard accuracy for BoolQ and WinoGrande, normalized accuracy for ARC and HellaSwag, and the macro average across all five tasks as the aggregate representation fidelity metric.

\subsection{Quality and Efficiency}
\paragraph{Accuracy.} Table~\ref{tab:quality_main} lists zero-shot accuracy across five benchmarks. We compare PCoMoE with MoE-I$^2$~\cite{yang2024moei2}, MoE-Pruner~\cite{xie2024moepruner}, and MoEITS~\cite{balderas2026moeits}. On Qwen1.5-MoE, PCoMoE improves the macro average score from 67.90 to 73.70, corresponding to a +5.80-point gain over Vanilla. Under the matched fine-tuning setting, PCoMoE further outperforms Vanilla-FT by +2.87 points. For Mixtral-8x7B and DeepSeek-V2-Lite, PCoMoE improves over Vanilla by +2.10 and +2.46 points, respectively, while retaining +1.61 and +2.14-point gains over their matched Vanilla-FT baselines. These results indicate that the activated off-diagonal paths introduce viable execution trajectories beyond the gains from standard fine-tuning without disrupting representation capacity.

\paragraph{Inference Acceleration.} Figure~\ref{fig:main_speed} details stage-specific token throughput. PCoMoE yields up to a 1.305$\times$ decode speedup and a 1.294$\times$ end-to-end speedup over vanilla baselines. On Mixtral-8x7B, decoding speed increases from 26.50 to 34.58 tokens per second, and end-to-end generation increases from 133.43 to 172.72 tokens per second. PCoMoE delivers over 20\% throughput improvements on Qwen1.5-MoE and maintains efficiency gains under the top-6 routing of DeepSeek-V2-Lite. Prefill performance remains identical to vanilla models because the layer-static masking and source-grouped compute reuse loop strictly target the autoregressive decoding phase.

\subsection{Ablation and Sensitivity Study}
\label{sec:ablation_sensitivity}

\paragraph{Design Ablation.}
Table~\ref{tab:design_ablation} isolates the effects of matched fine-tuning, router adaptation, path composition, and learned compatibility on Qwen1.5-MoE. Vanilla-FT improves the average accuracy from 67.90 to 70.83, showing that standard adaptation contributes part of the quality gain. However, Router-Only FT does not reproduce the improvement of PCoMoE, while composition with frozen compatibility biases increases Decode throughput but substantially degrades accuracy to 55.50. Full PCoMoE further improves over Vanilla-FT by 2.87 points while retaining a 1.238$\times$ Decode speedup, indicating that learned path compatibility is critical to the final quality--efficiency balance.

\begin{table}[t]
\setlength{\abovecaptionskip}{3pt}
\setlength{\belowcaptionskip}{3pt}
    \centering
    \footnotesize
    \caption{Design ablation on Qwen1.5-MoE. Decode throughput is measured in tokens/s.}
    \label{tab:design_ablation}
    \setlength{\tabcolsep}{2.5pt}
    \renewcommand{\arraystretch}{0.95}
    \begin{tabular*}{\columnwidth}{@{\extracolsep{\fill}}lccc@{}}
        \toprule
        \textbf{Method} &
        \textbf{Avg. Acc. (\%)} &
        \textbf{Decode} &
        \textbf{Speedup} \\
        \midrule
        Vanilla & 67.90 & 31.42 & 1.000$\times$ \\
        Vanilla-FT & 70.83 & 31.71 & 1.009$\times$ \\
        Router-Only FT & 67.51 & 32.37 & 1.030$\times$ \\
        Frozen-Bias & 55.50 & 40.41 & 1.286$\times$ \\
        \textbf{PCoMoE} & \textbf{73.70} & 38.90 & 1.238$\times$ \\
        \bottomrule
    \end{tabular*}
\end{table}

\begin{table}[t]
\setlength{\abovecaptionskip}{3pt}
\setlength{\belowcaptionskip}{3pt}
    \centering
    \footnotesize
    \caption{Ablation of PCoMoE execution optimizations on Qwen1.5-MoE. Throughput is measured in tokens/s.}
    \label{tab:system_ablation}
    \begin{tabular*}{\columnwidth}{@{\extracolsep{\fill}}lcc@{}}
        \toprule
        \textbf{Configuration} & \textbf{Throughput} & \textbf{Speedup} \\
        \midrule
        Base PCoMoE & 110.7 & 1.00$\times$ \\
        + Routing & 127.3 & 1.15$\times$ \\
        + Fusion & 141.7 & 1.28$\times$ \\
        + Dispatch & 160.5 & 1.45$\times$ \\
        + Routing + Dispatch & 171.6 & 1.55$\times$ \\
        + Fusion + Dispatch & 179.4 & 1.62$\times$ \\
        \textbf{All Optimizations} & \textbf{191.4} & \textbf{1.73$\times$} \\
        \bottomrule
    \end{tabular*}
\end{table}

\paragraph{Execution Ablation.}
Table~\ref{tab:system_ablation} isolates the end-to-end throughput impacts of PCoMoE execution optimizations on Qwen1.5-MoE. Base PCoMoE denotes the path-compositional model after expert decoupling and path construction, but before enabling the execution optimizations in Table~\ref{tab:system_ablation}. Path routing, fusion, and source-grouped dispatch individually yield 1.15$\times$, 1.28$\times$, and 1.45$\times$ speedups over this reference. Combining these components compounds efficiency, culminating in a 1.73$\times$ cumulative acceleration (191.4 tokens/s). Thus, hardware-aligned scheduling is required to translate path expansion into throughput gains.

\begin{figure}[t]
\setlength{\abovecaptionskip}{3pt}
\setlength{\belowcaptionskip}{3pt}
    \centering
    \includegraphics[width=\columnwidth]{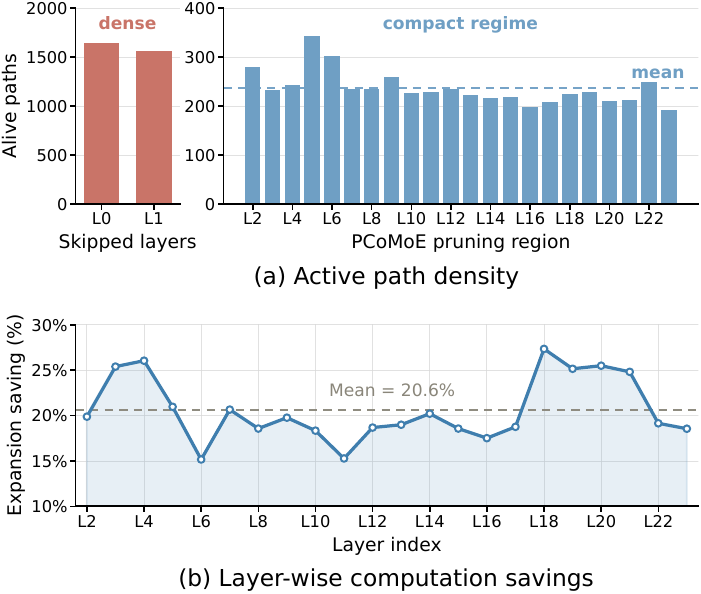}
    \caption{Layer-wise sensitivity analysis.}
    \label{fig:ablation_sensitivity}
    \vspace{-0.5cm}
\end{figure}

\paragraph{Pruning Sensitivity.} 
Figure~\ref{fig:ablation_sensitivity}(a) tracks active path density across layers. The initial two MoE blocks maintain high density because inseparable compatibility biases render early sparsification lossy. From the third block onward, active paths contract sharply due to the high separability of redundant routes. PCoMoE thus preserves the native configuration in the first two layers and executes structural pruning exclusively from the third block onward.

\paragraph{Reuse Heterogeneity.} 
Figure~\ref{fig:ablation_sensitivity}(b) shows the layer-wise arithmetic savings achieved by expansion-side reuse on Qwen1.5-MoE. Reductions average 20.6\% but vary from 15\% to 27\% across blocks, confirming that efficiency gains are non-uniform across the hierarchy. PCoMoE leverages this variance to dynamically allocate layer-wise path budgets based on local compatibility scores, improving overall throughput while avoiding excessive pruning in sensitive regions.

\section{Conclusion}
\label{sec:conclusion}

This paper introduces \textbf{PCoMoE}, an execution framework that shifts MoE inference from monolithic expert selection to fine-grained sub-transformation composition. By integrating structural path decomposition, compatibility-guided gating, and source-grouped compute reuse, PCoMoE leverages intra-expert computational redundancy while strictly maintaining native MoE layer interfaces. Evaluations show that PCoMoE improves both inference throughput and downstream representation fidelity over vanilla MoE baselines.

\section*{Acknowledgments}

This work was partially supported by the National Key Research and Development Program of China (2024YFE0204300), the National Natural Science Foundation of China (Grant No. 62402311), the Natural Science Foundation of Shanghai (Grant No. 24ZR1433700), and the Key Research and Development Program of Shanghai (25LN3201200).

\section*{Limitations}
First, while we demonstrate PCoMoE using standard SwiGLU-style MoE structures common in prevailing open-source LLMs, the underlying principle of path composition is generic and can be extended to alternative expert internals by adapting the operator boundaries. 
Second, our current system primarily optimizes autoregressive decoding, the primary latency bottleneck in MoE serving; integrating PCoMoE with complementary distributed or prefill-stage optimizations represents a promising and orthogonal future direction. 
Finally, although the offline calibration stage incurs zero inference-time overhead, further incorporating this process into automated end-to-end model compilation pipelines will help streamline broader deployments.



\bibliography{main/bibfile}

\end{document}